\documentclass[runningheads]{llncs}
\usepackage{graphicx}
\usepackage{algorithm}
\usepackage[noend]{algpseudocode}
\usepackage{amsmath}
\begin{document}
\title{Adversarially Robust Neural Legal Judgement Systems}
%
%
\author{Rohit Raj\inst{1}\orcidID{0000-0003-1421-6286} \and
V Susheela Devi\inst{1}
}
\authorrunning{R. Raj et al.}
%
\institute{Dept. of CSA, Indian Institute of Science,
Bangalore, Karnataka, India 
\email{\{rohitr,susheela\}@iisc.ac.in}}
\maketitle              
\begin{abstract}
Legal judgment prediction is the task of predicting the out-
come of court cases on a given text description of facts of cases. These
tasks apply Natural Language Processing (NLP) techniques to predict legal judgment results based on facts. Recently, large-scale public datasets
and NLP models have increased research in areas related to legal
judgment prediction systems. For such systems to be practically helpful, they should be robust from adversarial attacks. Previous
works mainly focus on making a neural legal judgement system; however, significantly less or no attention has been given to creating a robust Legal
Judgement Prediction(LJP) system. We implemented adversarial attacks
on early existing LJP systems and found that none of them could handle
attacks. In this work, we proposed an approach for making robust LJP
systems. Extensive experiments on four legal datasets show significant
improvements in our approach over the state-of-the-art LJP system in
handling adversarial attacks. To the best of our knowledge, we are the
first to increase the robustness of early-existing LJP systems
 
\end{abstract}
\section{Introduction}

Legal information is mainly in the form of text, so legal text processing is a growing area of research in NLP, such as crime classification \cite{a10}, judgment prediction \cite{a3}, and summarization \cite{a6}. Countries like India, which are highly populated, have many pending legal cases (approx 41 million ). In Brazil, only in the financial domain, three hundred thirty-two thousand cases are in progress \cite{a20}. It is due to multiple factors, including the unavailability of judges. Here legal judgment prediction system can help in several steps like finding articles or the history of a case, deciding penalty terms, etc. Also, legal judgment prediction is critical, so a small error in the system may drastically affect judicial fairness.\\
Most of the researchers focused on making LJP systems by training NLP models(LSTM, BERT \cite{a1}, legal-BERT \cite{a2}) on legal datasets. At the same time, very little or no attention has been given to the robustness of these models.\\

We summarise our contribution as follows:\\
1. We implemented adversarial attacks on existing baseline models after fine-tuning them on legal datasets and found that their performance decreased drastically.\\
2. We suggested an algorithm for adversarial training for making robust legal models.\\
3. We implemented training using data augmentation and adversarial training methods to improve the model's robustness.\\

\section{Related Work}

\subsection{Legal Judgement System}
Earlier legal judgment prediction systems involved linear models like SVM with a bag of words as feature representation. In recent years, neural network \cite{a3} methods have been used for legal domains due to the availability of NLP models like RNN and BERT \cite{a1}.\\
Most researchers used BiGRU-att \cite{a3}, HAN \cite{a3}, BERT \cite{a1} and Hier-BERT \cite{a4} architecture to predict article violation on ECtHR \cite{a5} dataset. Legal-BERT \cite{a2} is a domain-specific BERT pretrained on legal-documents corpora of approx 11.5 GB, used for legal judgement prediction. A number of other tasks like legal summarization \cite{a6},prior case retrieval \cite{a8},legal QA \cite{a7} have been introduces.\\
In legal judgment prediction, the model must predict the final decision based on case facts. Several datasets are introduced for training so that model can learn specific words (for example, ‘restrictive covenant’, ‘promissory estoppel’, ‘tort’ and ‘novation’) that are being used in legal documents which are not used for general purposes; for example, ECtHR \cite{a5}, a multilabel dataset containing violated articles as the label. SCOTUS \cite{a5} contains cases of the American Supreme Court and ILDC \cite{a9} contains cases of the Indian supreme court. All of these are English datasets. However, datasets from different languages are also introduced like Chinese \cite{a10}, Swedes \cite{a11}, Vietnamese \cite{a12}.\\

\subsection{Adversarial Training}
Several adversarial training methods have been explored in NLP models to increase their robustness. The models are trained on a dataset containing augmented adversarial examples with the original dataset in adversarial training. These adversarial examples are generated by applying adversarial attacks on pre-trained models such that generated examples should be similar to the original example, and the average human user cannot differentiate it from the natural one. Several adversarial attack mechanisms are being used in NLP, such as BERT-Attack \cite{a15}, BAE \cite{a14}, A2T \cite{a16}, TextFooler \cite{a13}. In these attacks, the model finds essential words in the original text and replaces them with semantically similar words such that the label of the original text changes and generates adversarial text that looks similar to the original text.

\subsection{Why adversarial training ?}
To motivate the necessity of adversarial training, we implemented adversarial attacks
on existing baseline models(BERT \cite{a1}, Legal-BERT \cite{a2}, RoBERTa \cite{a18}) to check their
robustness. We found that the performance of these models decreased drastically, as these models could not handle the adversarial attack. We also implemented data augmentation using back-translation during training, but the model's performance was not improved much.\\
Legal judgment prediction is critical, so a slight variation in the input may affect judgment fairness. So during deployment, if someone intentionally perturbs the input sequence, prediction may change drastically. It is
the main reason for adversarial training.\\

\section{Problem Formulation}
Given a legal dataset, which contains a collection of legal documents, $L = \{(X_1,y_1)..(X_N,y_N)\}$, where $X_i$ is a legal text extracted from a legal document and $y_i = \{1,2,3..K\}$. Here the length of each $X_i$ is very large, and $y_i$ is a label corresponding to that text. \\
The task is to design an LJP model $M(.)$ that can:
\begin{itemize}
    \item Predict correct class on legal documents of even large length.
    \item Perform correct prediction even if data is perturbed. Let $X'$ be a perturbed text, which may be perturbed intentionally or by mistake, then $M(X') \rightarrow y$, where $y$ is the correct label of that legal text.
\end{itemize}

\section{Methods}
 In this section, we present our training workflow. We implemented three methods for training. These are 1) Fine-tuning Baseline models, 2) Training baseline models with data augmentation 3) Adversarial training using augmenting adversarial examples with natural examples. At the end of each method, we tested our model's robustness with adversarial attacks.
 \subsection{Fine tuning baseline model}
 In this approach, we have taken baseline models( BERT \cite{a1}, Legal-BERT \cite{a2}, RoBERTa \cite{a18}, Hierarchical Version of BERT \cite{a4}, we have used a modified version of Hier-BERT, denoting as H-BERT) and fine-tuned them on our downstream tasks for legal judgment predictions. For BERT, Legal-BERT, and RoBERTa, we have taken the last 512 tokens of each input text for training, as this approach gave a better result. For H-BERT(modified Hierarchical Version of BERT), we have divided text into chunks of 510 tokens such that two consecutive chunks overlapped each other, here RoBERTa is taken as encoder, shown in Figure 2, as it gives the best result. We have used cross-entropy as a loss function for updating the gradient and evaluated model performance on accuracy.\\
 \subsection{Training using data-augmentation}
 In this approach, we first generated data using back-translation\cite{a24} and then augmented it with training data. The algorithm for training is similar to Algorithm 2, where in place of an adversarial example generator, we are using a data augmenter.\\
 We use the transformer model implemented by HuggingFace \cite{a21} for back-translation. We first translate English to French and then translate it back to English from French. We augment newly generated data such that it does not have any duplicate instances, and training is done similarly to approach 1.\\
 
 \subsection{Adversarial Training}
 In this approach, we generate adversarial examples from original legal document datasets, then further augment these examples with legal document datasets and train the model on this new dataset, i.e., $ D_{new} = D_{nat} \cup D_{adv}$.\\
 For generating an adversarial example from a text sample, first, we find the importance score of each word in that sample using greedy search with word importance ranking mechanism \cite{a25}, where the importance of the word is determined by how much heuristic score changes when a word is deleted from the original input. i.e.,\\
     
\begin{equation}
I_{w_{i}} = 
\left\{
\begin{array}{ll}
 M_{y}(X) - M_{y}(X/_{w_{i}}) , & \mbox{if $ M(X) = M(X/_{w_{i}}) = y$},\\
(M_{y}(X) - M_{y}(X/_{w_{i}})) + (M_{y}^{'}(X/_{w_{i}}) - M_{y}^{'}(X)), & \mbox{if $ M(X) = y, M(X/_{w_{i}}) = y^{'} $ } \\
& \mbox{and $y \neq y^{'} $} \\

& \mbox{\label{eq1}}
\end{array}
\right.
\end{equation}
\\
Here we have followed the deletion approach for finding word importance because we are considering a common black-box setup which is usually followed in a real-world scenario. We denote sentence after deletion of word $w_{i}$ as $X/_{w_{i}}$ = $\{w_{1},...,w_{i_{-_{1}}}, w_{i_{+_{1}}},...w_{n}\}$ and use $M_{y}(.)$ to denote prediction score of model for label $y$. Here $I_{w_{i}}$ denote importance score of word $w_{i}$ which is defined in equation 1.\\
As shown in Algorithm 1, in lines 3-4, we select top-k words according to importance score using equation 1 and generate `m' synonyms for each word using cosine-similarity and counter-fitted-word-embedding \cite{a23}. We then replace original words with synonyms and make an adversarial example $X'$. Further, to find the similarity of an adversarial sample $X'$ to the original sample $X$, we use Universal Sentence Encoder(USE) \cite{a19}. We ignore the examples below a certain threshold value. We have taken 0.5 as the threshold value for all of our experiments. We have implemented all of our algorithms on top of the Textattack framework.

\begin{algorithm}
\caption{Adversarial Example Generation from legal Sample}\label{alg:euclid}
\begin{algorithmic}[1]
\State \textbf{Input:} {Legal judgement prediction model $M(\theta)$, legal sample sentence $X = (w_{1},w_{2},..w_{n})$, Perturbation Generator $P(X,i)$ which replace $w_i$ with certain perturbed word using counter-fitted-word-embedding }
\State \textbf{Output:} {Adversarial legal sample $X_{adv}$}
\State Calculate importance score $I(w_i)$ of each word $w_i$ using equation 1.
\State Take top-k words and rank them in decreasing order according to $I(w_i) $ and store them in set $R = (r_{1},r_{2}..r_{k})$ 
\State $X' \leftarrow X$ 
\State \textbf{for $i = r_{1},..r_{n}$ in $R$ do,}
\State \hspace{0.5cm}$X_p \leftarrow$ perturb the sentence $X'$ using $P(X',i)$
\State \hspace{0.5cm} \textbf{if $M(X_p) \neq y$ then }
\State \hspace{1.0cm} \textbf{if $sim(X_p, X) > threshold$ then}
\Comment{Check similarity of $X$ and $X'$}
\State \hspace{1.5cm}       $X' \leftarrow X_p$
\State \hspace{1.0cm} \textbf{end if}
\State \hspace{0.5cm} \textbf{end if}
\State \textbf{end for}
\State \textbf{return $X'$ as $X_{adv}$ }

\end{algorithmic}
\end{algorithm}

For adversarial training, we first train the model using natural legal dataset $D_{nat}$ for some iterations,i.e., $n_{nat}$, after that we augment adversarial example $D_{adv}$ by using adversarial example generator, i.e., $D_{new} = D_{nat} \cup D_{adv}$. Further, train the model on $D_{new}$ for some iterations, i.e., $n_{adv}$. Here adversarial loss function is used to train the model.\\
Let $L_{nat}$ be the loss function used for natural training, which is defined as a cross-entropy loss function, i.e.,
 \begin{equation}
L_{nat} = L_{\theta}( X, y)
\end{equation}
Where $X$ is the input text and $y$ is the label corresponding to it.
If $A_\theta(X,y)$ is the adversarial example generator, then the loss function for adversarial training is defined as a cross-entropy loss function,i.e
 \begin{equation}
L_{adv} = L_{\theta}(A_{\theta}(X,y), y)
\end{equation}
So our final loss function will be the combination of these two cross entropy loss functions, i.e.,
\begin{equation}
  L = argmin_{\theta} (L_{nat} + \gamma L_{adv})
\end{equation}
Where $\gamma$ is a hyper-parameter that is used to change the importance of adversarial training.\\
As shown in Algorithm 2, lines 3-6 represent the natural training of the model. Lines 7-18 represent the adversarial training of the model, which is pre-trained in lines 3-6. In lines 10-15, adversarial examples are generated. Line 16 represents the augmentation of adversarial examples with natural data. Line 17 represents the adversarial training step.\\

\begin{algorithm}
\caption{Adversarial Training of legal Models}\label{alg:algo 2}
\begin{algorithmic}[1]
\State \textbf{Input:} {Legal judgement prediction model $M_{\theta}(.)$, Adversarial example generator algorithm $A_{\theta}(X,y)$, legal dataset $D_{nat} = \{X,y\}_{i_{=_{1}}}^{m}$, natural training epochs $n_{nat}$, adversarial training epochs $n_{adv}$ }
\State \textbf{Output:} {Adversarially trained model }
\State Randomly initialize $\theta$
\State \textbf{for $i = 1,2.. n_{nat}$ do,}
\State \hspace{0.5cm} Train  $M_\theta(.)$ on dataset $D_{nat}$ using loss function from Equation (2).
\State \textbf{end for}
\State \textbf{for $i = 1,2.. n_{adv}$ do,}
\State \hspace{0.5cm} Initialize set of adversarial legal dataset $D_{adv} \leftarrow \{\}$
\State \hspace{0.5cm} $K \leftarrow $ fraction of adversarial samples to be generated of natural dataset
\State \hspace{0.5cm} \textbf{for i = 1,2..$size(D_{nat} )$ do,}
\State \hspace{1.0cm}\textbf{if size($D_{adv}$)$ < K * D_{nat}$ then}
\State \hspace{1.5cm} $X_{adv} \leftarrow A_\theta(X,y)$ 
\State \hspace{1.5cm} $D_{adv} \leftarrow D_{adv} \cup \{X_{adv}, y\}$
\State \hspace{1.0cm}\textbf{end if}
\State \hspace{0.5cm}\textbf{end for}
\State \hspace{0.5cm} $D_{new} \leftarrow D_{nat}  \cup D_{adv}$
\State \hspace{0.5cm} Train $M_\theta(.)$ on $D_{new}$ using loss function from Equation (4).
\State \textbf{end for}
\end{algorithmic}
\end{algorithm}

\section{Experiments and Results}
\subsection{Datasets and Models}
\subsubsection{Datasets}
\paragraph{}
\textbf{ECHR \cite{a3}:} It contains cases of the European Council of Human Rights (ECHR). The dataset has 11.5k cases, of which  7100 cases are used for training, 1380  for development, and 2998 for the test set. The training and development set contains cases from 1959-2013, and the test set contains cases from 2014-2018. Total ECHR articles are 66; however, we have taken \textbf{binary representation} of the ECHR dataset, in which label 1 is assigned if any article is violated; otherwise, 0 is assigned. \\\\
\textbf{SCOTUS\cite{a5} :} It is a dataset of the US Supreme Court, which hears only complex cases not well solved by lower courts. SCOTUS is a multi-class single label dataset containing 14 classes, broad areas like Civil Rights, Criminal Procedure, Economic Activity, etc. The SCOTUS cases are split into a 5k(1946-1982) training set, 1.4k(1982-1991) development set, and 1.4k(1991-2016) test set.\\
\textbf{ILDC : }Indian Legal Document Corpus (ILDC) is introduced by \cite{a9}, which contains cases of the Supreme court of India(SCI) from 1947 to 2020. It is a \textbf{binary classification} dataset contain labels $\{0,1\}$. It has two versions. \textbf{1)ILDC-single} contains cases of a single petition filed, label 1 is assigned to cases whose petition is accepted, and 0 is for not accepted. \textbf{2)ILDC-multi} contains cases with multiple petitions filed. Here label 1 is assigned to cases with at least one petition accepted; otherwise, label 0 is assigned. We have taken ILDC-multi for all of our experiments.\\
As shown in Figure 1, legal text datasets have a substantial length. Here the average length of samples of ECtHR is \textbf{1619} words,  SCOTUS is \textbf{5853} words, and ILDC is \textbf{3208} words. So it is far greater than a normal BERT architecture input size. Therefore, we have implemented the modified Hierarchical Variant of BERT (H-BERT) architecture.\\
\begin{figure}
\includegraphics[width=\textwidth, height=3cm]{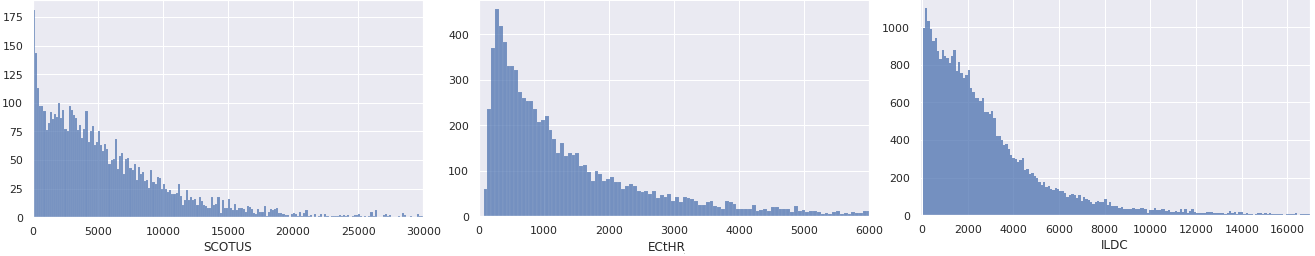}
\caption{Text length distribution of different datasets, here horizontal axis shows length of input texts and vertical axis show number of inputs} \label{fig1}
\end{figure}

\subsubsection{Models}
\paragraph{}
\textbf{BERT}\cite{a1} is a pre-trained transformer-based language model. It is pre-trained to perform masked language modeling and next sentence prediction.\\
\textbf{Legal-BERT} \cite{a2} is BERT pre-trained on English legal corpora, which contains legislation, contracts, and court cases. Its configuration is the same as the original BERT configuration. The sub-word vocabulary of Legal-BERT is built from scratch.\\
\textbf{Hierarchical Variant of BERT (H-BERT)} Legal documents are usually of large text length (shown in Figure 1), for example, ECtHR, ILDC, and SCOTUS. Transformer-based models can handle up to 512 sub-word units. So we implemented an architecture similar to \cite{a4} in which we divided the text into the chunk of 510 tokens such that two consecutive chunks have 100 overlapping tokens. Each chunk is sent through a BERT-Encoder to generate CLS embedding. As shown in Figure 2, CLS embedding is passed to 1-dimensional convolution and max-pooling layers. A further output of the max-pooling layer is passed to Bi-directional LSTM and then the Dense layer. We have taken RoBERTa as an encoder here as it gave best result among all other BERT-based models.\\

\subsection{Implementation Details}
For all tasks, we use pre-trained transformer-based BERT models from Huggingface  implementation. Each model output a 768-dimension vector regarding each input text. The batch size is set to 8. Models trained using Adam optimizer with 1e-5 learning rate for overall 10 epochs, which includes 3 epochs of natural training and 7 epochs of adversarial training. We used LSTM of 100 units and 1-D CNN with 32 filters for H-BERT.

\begin{figure}
\includegraphics[width=\textwidth, height=8cm]{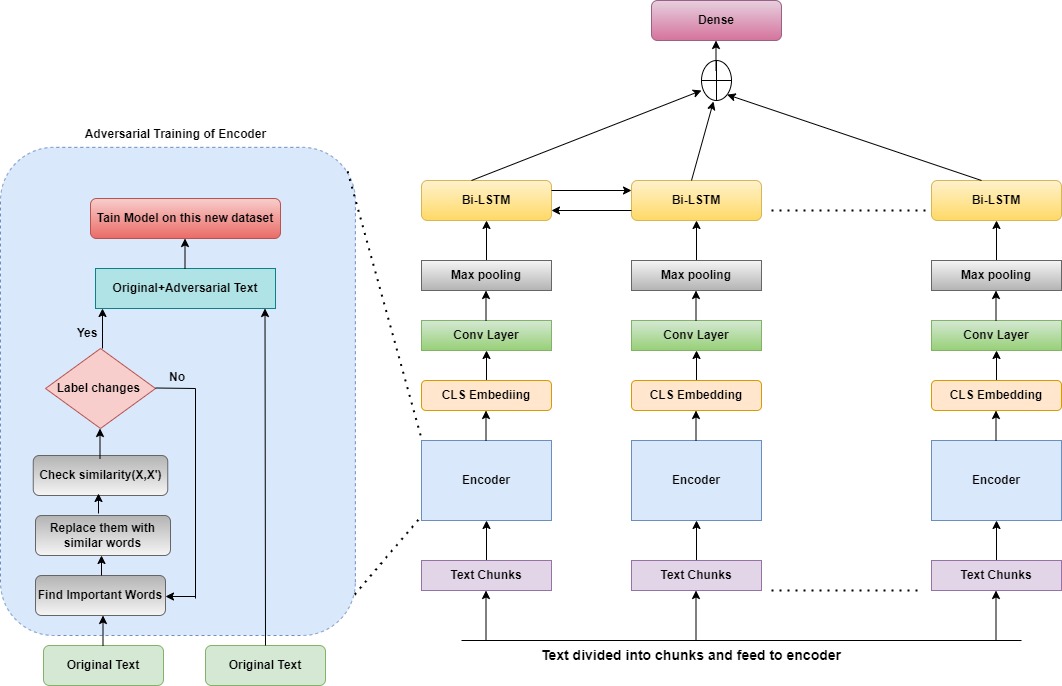}
\caption{Adversarially Robust Neural Legal Judgement Model (H-BERT model)} \label{fig1}
\end{figure}

\subsection{Results}
\subsubsection{Results after fine tuning}
\paragraph{}
We have fine-tuned models naturally, i.e., without any augmentation (shown in Table \ref{acc_nat_tr}). For BERT, Legal-BERT, and RoBERTa, we have taken the last 512 tokens of each sample as input, and for H-BERT, we have divided the text into chunks, as mentioned in section \ref{data_model_section}. From empirical results, we can say that H-BERT performs better than other models as H-BERT takes whole text examples, whereas other models take only the last 512 tokens. Legal-BERT performs better on ECHR and SCOTUS datasets as it is pre-trained on legal documents of Europe and America. The performance of RoBERTa on the ILDC dataset is better than Legal-BERT because Legal-BERT is not pre-trained on the Indian-origin legal dataset, whereas  RoBERTa is pre-trained on general English datasets.\\

\begin{table}
\centering
\caption{Accuracy of Naturally trained models, $(FT)$ : Fine Tuning}\label{acc_nat_tr}
\begin{tabular}{|l|l|l|l|l|}
\hline
\textbf{Models} &  \textbf{ECHR} & \textbf{SCOTUS}  & \textbf{ILDC$_{multi}$} \\
\hline
BERT$(FT)$ & 81.21  & 68.33  & 67.24 \\
Legal-BERT$(FT)$  & 83.42  & 76.47 & 63.37 \\
RoBERTa$(FT)$  & 79.27 & 71.69  & 71.26\\
 H-BERT$(FT)$  & 81.03  & 78.02  & 74.89\\
\hline
\end{tabular}
\end{table}

\subsubsection{Results after adversarial attack on naturally trained models}
\paragraph{}
We feed 1000 adversarial examples generated from the adversarial examples generator to naturally trained models to check their robustness against adversarial attacks. As shown in Table \ref{acc_actk_nat}, naturally trained models could not handle adversarial attacks as their accuracy decreased drastically. The accuracy of BERT decreased the most because it is not pre-trained on domain-specific (legal domain) datasets, whereas, in the case of H-BERT, accuracy decreased least because H-BERT's RoBERTa  is pre-trained on general English datasets as well as during training, it is considering whole legal text documents. In contrast, other models consider only the last 512 words of each example. Legal-BERT is more robust than BERT as it is pre-trained on legal datasets. RoBERTa is pre-trained on a large corpus, so it can able to handle adversarial attacks better than Legal-BERT.\\

\begin{table}
\caption{Accuracy of Naturally trained Models after attack, $(FT)$ : Fine Tuning}\label{acc_actk_nat}
\begin{tabular}{|l|l|l|l|l|}
\hline
\textbf{Models} &  \textbf{ECHR} & \textbf{SCOTUS}  & \textbf{ILDC$_{multi}$} \\
\hline
BERT$(FT)$ &  33.12  &  36.42 & 22.59 \\
Legal-BERT$(FT)$  & 36.27  & 41.67  & 25.26 \\
RoBERTa$(FT)$  & 36.05  & 41.91  & 38.92\\
H-BERT$(FT)$  & 39.18  & 43.19  & 37.21\\
\hline
\end{tabular}
\end{table}

\subsubsection{Results after adversarial attack on model trained using data-augmentation}
\paragraph{}
We feed 1000 adversarial examples to a model trained using data augmentation to check their robustness. As shown in Table \ref{acc_atck_da}, the accuracy of models is less than that of naturally trained models but better than the accuracy of models after the adversarial attack on naturally trained models. This is because we are augmenting extra data, which is very similar to the original data except for a few words for training. So due to this, the model is more diverse and can handle some adversarial attacks. In most cases, H-BERT performs better than others because it considers whole text data instead of the last 512 tokens. \\

\begin{table}
\caption{Accuracy after adversarial attack , $(DA)$ : Data Augmentation }\label{acc_atck_da}
\begin{tabular}{|l|l|l|l|l|}
\hline
\textbf{Models} &  \textbf{ECHR}  & \textbf{SCOTUS} & \textbf{ILDC$_{multi}$} \\
\hline 
BERT$(DA)$ \textbf{(Ours)}   & 38.03  & 41.12 & 32.56 \\
Legal-BERT$(DA)$ \textbf{(Ours)} & 39.36  & 43.15 & 41.66 \\
RoBERTa$(DA)$ \textbf{(Ours)}  & 40.21  & \textbf{45.09}  & 38.71\\
H-BERT$(DA)$ \textbf{(Ours)}  & \textbf{46.10}  & 45.02  & \textbf{42.03}\\
\hline
\end{tabular}
\end{table}

\subsubsection{Results after adversarial training}
\begin{figure}
\includegraphics[width=14cm, height=3cm]{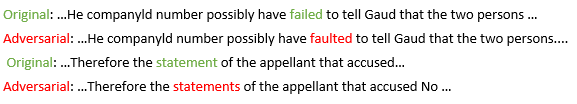}
\caption{original and adversarial examples of ILDC dataset while training BERT.} \label{ildc_adv}
\end{figure}

\paragraph{}
We implemented adversarial training using our Algorithm 2. As shown in Table \ref{acc_aftr_adv_tr}, sometimes, the accuracy of an adversarially trained model is better than the naturally trained model. The increase in accuracy is due to the augmentation of adversarial examples, which creates more diversity during training. The performance of the H-BERT model is best, while Legal-BERT is performing better on ECHR and the SCOTUS dataset because it is pretrained on European and American legal documents. Figure \ref{ildc_adv} shows an adversarial example on the ILDC dataset during adversarial training. As we can see, slight change perturbation in the text can change the label of an input. Due to the large length of text input, we have shown only a small snippet of an example where an example is being perturbed. \\

\begin{table}
\caption{Accuracy after adversarial training, $(AT)$ : Adversarial Training}\label{acc_aftr_adv_tr}
\begin{tabular}{|l|l|l|l|l|}
\hline
\textbf{Models} &  \textbf{ECHR}  & \textbf{SCOTUS} & \textbf{ILDC$_{multi}$} \\
\hline
BERT$(AT)$ \textbf{(Ours)} &  79.23  & 69.07 & 65.56 \\
Legal-BERT$(AT)$ \textbf{(Ours)}  & 82.01   & 77.02 & 61.02 \\
RoBERTa$(AT)$ \textbf{(Ours)} & 81.73  & 70.03  & 69.97\\
H-BERT$(AT)$ \textbf{(Ours)} & \textbf{83.67}  & \textbf{78.09}  & \textbf{71.53}\\
\hline
\end{tabular}
\end{table}

\subsubsection{Results after adversarial attack on adversarially trained model}
\paragraph{}
We feed 1000 adversarial examples, as earlier, to check the robustness of the adversarially trained model. The results are surprising, as shown in Table \ref{acc_aftr_attack_at}. Our models can handle most adversarial attacks. Accuracy is far better than accuracy after the attack on naturally trained models. This is because, during adversarial training, the model came across a diverse set of words that were not present earlier.\\
As shown in Table \ref{acc_aftr_attack_at}, H-BERT is performing better than other models because it is trained on the whole dataset. The BERT model performs worst as it is not pre-trained on legal documents. The performance of Legal-BERT is not satisfactory on ILDC because it is pre-trained on European and American legal documents, which may contain words that are different from Indian legal documents.

\begin{table}
\centering
\caption{Accuracy after attack , $(AT)$ : Adversarial Training}\label{acc_aftr_attack_at}
\begin{tabular}{|l|l|l|l|l|}
\hline
\textbf{Models} &  \textbf{ECHR}  & \textbf{SCOTUS} & \textbf{ILDC$_{multi}$} \\
\hline
BERT$(AT)$ \textbf{(Ours)} &  58.96  & 52.38  & 54.46 \\
Legal-BERT$(AT)$ \textbf{(Ours)}  & 64.07  & 52.71  & 51.96 \\
RoBERTa$(AT)$ \textbf{(Ours)}  & 64.97  & 50.09  & 55.91\\
H-BERT$(AT)$ \textbf{(Ours)}  & \textbf{69.32}  & \textbf{61.53}  & \textbf{58.29}\\
\hline
\end{tabular}
\end{table}

\section{Conclusion and Future work}
In this work, we empirically proved that early existing legal models are not adversarially robust, which is a significant risk for deploying them in work. We also presented an adversarially robust model, which is trained on our adversarial training algorithm for legal judgment prediction, which performs better than state-of-the-art models in the presence of adversarial examples.\\
For future work, we suggest making robust legal models which can be applied to Legal documents that are different from English. Also, one can work on zero-shot and few-shot learning in legal domains, where very few resources are available for legal documents.

%
%
%

\begin{thebibliography}{8}
\bibitem{a1}
Devlin,J., Chang,M., Lee,K., Toutanova,k. 2019 BERT: Pre-training of Deep Bidirectional Transformers for Language Understanding
https://doi.org/10.48550/arXiv.1810.04805
\bibitem{a2}
Chalkidis, I, M Fergadiotis, P Malakasiotis, N Aletras, I Androutsopoulos 2019 LEGAL-BERT: The Muppets straight out of Law School 
https://doi.org/10.48550/arXiv.2010.02559
\bibitem{a3}
I Chalkidis, I Androutsopoulos, N Aletras 2019
Neural Legal Judgment Prediction in English
https://doi.org/10.48550/arXiv.1906.02059
\bibitem{a4}
I Chalkidis, M Fergadiotis, D Tsarapatsanis, N Aletras, I Androutsopoulos, Ps Malakasiotis 2021
Paragraph-level Rationale Extraction through Regularization: A case study on European Court of Human Rights Cases
http://dx.doi.org/10.18653/v1/2021.naacl-main.22
\bibitem{a5}
I Chalkidis, A Jana, D Hartung, M Bommarito, I Androutsopoulos, D Martin Katz, N Aletras 2021
LexGLUE: A Benchmark Dataset for Legal Language Understanding in English
https://doi.org/10.48550/arXiv.2110.00976
\bibitem{a6}
Vu Tran, Minh Le Nguyen, and Ken Satoh. 2019
Building Legal Case Retrieval Systems with Lexical
Matching and Summarization Using A Pre-Trained
Phrase Scoring Model. In Proceedings of the Seventeenth International Conference on Artificial Intelligence and Law, ICAIL ’19, page 275–282, New
York, NY, USA.
\bibitem{a7}
H. Zhong, Y. Wang, C. Tu, T. Zhang, M. Sun, Iteratively questioning and answering for interpretable legal judgment prediction, Proceedings
of the AAAI Conference on Artificial Intelligence 34 (1) (2020) 1250–1257.

\bibitem{a8}
P Jackson, K Al-Kofahi, A Tyrrell, and
A Vachher. 2003. Information extraction from
case law and retrieval of prior cases. Artificial Intelligence, 150(1-2):239–290

\bibitem{a9}
V Malik, R Sanjay, S Kumar Nigam, K Ghosh, S Guha, A Bhattacharya, A Modi
ILDC for CJPE: Indian Legal Documents Corpus for Court Judgment Prediction and Explanation 2021
https://doi.org/10.48550/arXiv.2105.13562
\bibitem{a10}
C Xiao, H Zhong, Z Guo, C Tu, Z Liu, M Sun, Y Feng, X Han, Z Hu, H Wang, et al. 2018.
Cail2018: A large-scale legal dataset for judgment
prediction. arXiv preprint arXiv:1807.02478.
\bibitem{a11}
 J. Niklaus, I. Chalkidis, M. Stürmer, Swiss-judgment-prediction: A multilinguallegal judgment prediction benchmark, arXiv preprint
arXiv:2110.00806 (2021).
\bibitem{a12}
h Chau, T Nguyen, and Lh Nguyen. Vnlawbert: A vietnamese legal answer selection approach using bert language model. In 2020 7th NAFOSTED Conference on Information and Computer Science (NICS), pages 298–301. IEEE, 2020.
\bibitem{a13}
D Jin, Z Jin, J Zhou, P Szolovits Is BERT Really Robust? A Strong Baseline for Natural Language Attack on Text Classification and Entailment
https://doi.org/10.48550/arXiv.1907.11932
\bibitem{a14}
S Garg, G Ramakrishnan 2020
BAE: BERT-based Adversarial Examples for Text Classification
https://doi.org/10.48550/arXiv.2004.01970
\bibitem{a15}
Lg Li, R Ma, Q Guo, X Xue, X Qiu 2020
BERT-ATTACK: Adversarial Attack Against BERT Using BERT
https://doi.org/10.48550/arXiv.2004.09984
\bibitem{a16}
J Yoo, Y Qi 2021
Towards Improving Adversarial Training of NLP Models
https://doi.org/10.48550/arXiv.2109.00544
\bibitem{a17}
M Ribeiro, S Singh, C Guestrin 2016
"Why Should I Trust You?": Explaining the Predictions of Any Classifier
https://doi.org/10.48550/arXiv.1602.04938
\bibitem{a18}
Y Liu, M Ott, N Goyal, J Du, M Joshi, D Chen, O Levy, M Lewis, L Zettlemoyer, V Stoyanov
RoBERTa: A Robustly Optimized BERT Pretraining Approach
https://doi.org/10.48550/arXiv.1907.11692
\bibitem{a19}
D Cer, Yi Yang, S Kong, N Hua, N Limtiaco, Rhomni St. John, N Constant, M Guajardo-Cespedes, S Yuan, C Tar, Y Sung, B Strope, R Kurzweil
Universal Sentence Encoder
https://doi.org/10.48550/arXiv.1803.11175
\bibitem{a20}
L. Garcia-Navarro, P. MOURA, 2014 Brazil: The land of many lawyers and very slow justice .
https://www.npr.org/transcripts/359830235
\bibitem{a21}
https://huggingface.co/models
\bibitem{a22}
N Mrksic, D Séaghdha, B Thomson, M Gasic, L Rojas-Barahona, P Su, D Vandyke, T Wen, and
Steve J. Young. 2016. Counter-fitting word vectors
to linguistic constraints. In HLT-NAACL.
\bibitem{a23}
N Mrkšić, D  Séaghdha, BThomson, M Gašić, L Rojas-Barahona, P Su, DVandyke, T Wen, S Young 2016
Counter-fitting Word Vectors to Linguistic Constraints
https://doi.org/10.48550/arXiv.1603.00892
\bibitem{a24}
Q Xie, Z Dai, E Hovy, M Luong, Q V. Le 2019
Unsupervised Data Augmentation for Consistency Training
https://doi.org/10.48550/arXiv.1904.12848

\bibitem{a25}
J Yoo, J Morris, E Lifland, Y Qi
Searching for a Search Method: Benchmarking Search Algorithms for
Generating NLP Adversarial Examples
https://aclanthology.org/2020.blackboxnlp-1.30/
\end{thebibliography}
%

\end{document}